\documentclass[10pt,twocolumn,letterpaper]{article}

\usepackage[pagenumbers]{cvpr}  % To produce the arXiv version

\usepackage{cuted} %< for strip environment

\usepackage{currfile} %< for \currfilebase macro

\usepackage{caption} %< for "\captionof" commands (teaser)
\usepackage{xcolor}
\usepackage[most]{tcolorbox}

\usepackage{multirow}
\usepackage{placeins}
\usepackage{float}

\definecolor{cvprblue}{rgb}{0.21,0.49,0.74}
\usepackage[pagebackref,breaklinks,colorlinks,allcolors=cvprblue]{hyperref}

\def\confName{CVPR}
\def\confYear{2027}

\title{DiCT: Teaching X-ray Vision-Language Models to Reason via 
Imagining CT}

\author{%
    {\fontsize{10.95pt}{13.6pt}\selectfont
        Zhenghao Lu\textsuperscript{1\textdagger}\enspace
        Xiaotian Zhang\textsuperscript{1\textdagger}\enspace
        Songtao Jiang\textsuperscript{1}\enspace
        Yiqun Lin\textsuperscript{2}\enspace
        Siming Fu\textsuperscript{1}\enspace
        Hualiang Wang\textsuperscript{1*}\enspace
        Zuozhu Liu\textsuperscript{1*}}\\[0.4ex]
    {\normalsize
        \textsuperscript{1}Zhejiang University\qquad
        \textsuperscript{2}University College London}\\[0.4ex]
    {\small\texttt{\fontencoding{T1}\selectfont
        zhenghaolu@zju.edu.cn, 
        hualiang\_wang@zju.edu.cn, zuozhuliu@intl.zju.edu.cn}}%
}

\begin{document}
\maketitle
\iftoggle{cvprfinal}{%
    \begingroup
    \renewcommand{\thefootnote}{\fnsymbol{footnote}}%
    \footnotetext[2]{Equal contribution.}%
    \footnotetext[1]{Corresponding authors.}%
    \endgroup
}{}
\begin{abstract}
Diagnosing with X-rays often requires reasoning about the 3D locations of abnormalities and their spatial relationships to surrounding anatomical structures, yet such reasoning remains challenging due to the loss of spatial information in 2D projections. To address this challenge, we propose \textbf{D}reaming \textbf{i}n \textbf{CT} (\textbf{DiCT}), a three-stage progressive CT-to-X-ray reasoning transfer framework that leverages a CT vision-language model (VLM) to guide an X-ray VLM. Through this CT-guided transfer, the student acquires CT-informed 3D reasoning capabilities, enabling it to reason directly from 2D X-rays without explicit CT reconstruction at inference time. The framework begins with instruct the dreamer, which equips the X-ray student with initial CT-oriented reasoning capabilities and establishes a structured interface that distinguishes X-ray observations from CT imagination. Building on this structure, visualize the dream introduces CT-guided on-policy distillation (OPD), where the CT teacher selectively guides the X-ray student's reasoning trajectories to transfer 3D anatomical knowledge. Finally, manifest the dream adapts the acquired reasoning capabilities from digitally reconstructed radiographs (DRRs) to real clinical X-rays. Experiments across two VLM backbones on 3D-RAD and ReXVQA demonstrate DiCT's effectiveness across radiological reasoning tasks, achieving a 1.66 percentage point improvement over a CT reconstruction-based baseline on 3D-RAD Abnormality Identification with 12.77$\times$ lower inference computation.
\iftoggle{cvprfinal}{Our code is available at \url{https://github.com/HongYep/DiCT}.}{}
\end{abstract}

\section{Introduction}
\label{sec:intro}

\begin{figure}[t]
\centering
\includegraphics[width=\linewidth]{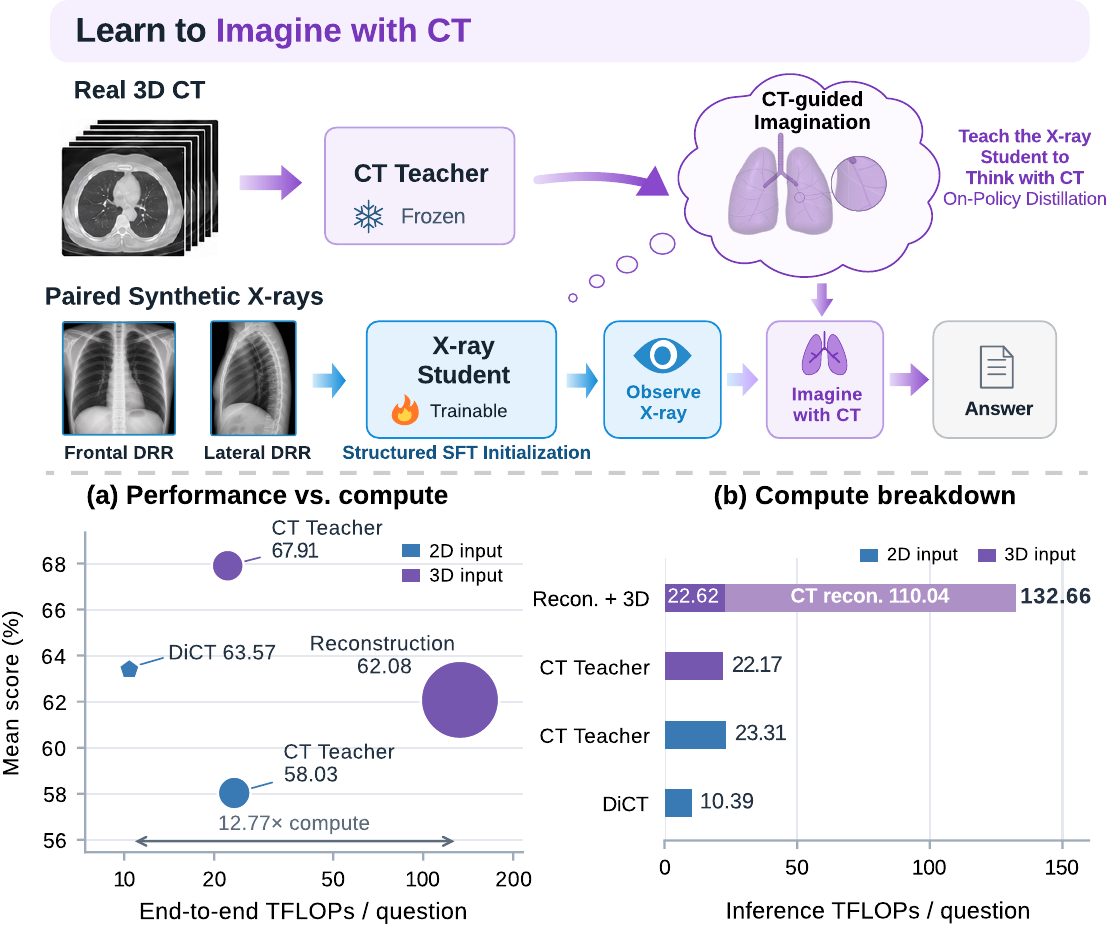}
\caption{\textbf{Overview and performance comparison of DiCT.} The upper panel illustrates how DiCT transfers CT-informed reasoning from a CT teacher to an X-ray student for X-ray-only inference without CT reconstruction. The lower panel compares model performance and inference cost across methods. Mean score is the arithmetic mean accuracy of Abnormality Identification and Longitudinal Diagnosis on 3D-RAD. $12.77\times$ denotes the reduction in inference FLOPs over the TF-PRDiT + ClinFusion-8B-SFT reconstruction pipeline.}
\label{fig:ct_imagination}
\end{figure}

X-ray radiography is widely used in clinical practice due to its rapid acquisition, broad accessibility, and relatively low radiation exposure. In contrast, computed tomography (CT) provides detailed 3D anatomical information, enabling precise localization of abnormalities and assessment of their spatial relationships to surrounding structures, but typically involves higher radiation exposure, longer acquisition times, and reduced accessibility. Given the accessibility of X-rays and the rich spatial information provided by CT, obtaining CT-informed 3D anatomical understanding from X-ray images alone would be highly valuable.

Recent advances in vision-language models (VLMs) have shown promising capabilities in radiological image understanding and question answering. However, reasoning about 3D anatomy from X-rays remains challenging for VLMs, as the projection of 3D anatomy onto 2D images results in overlapping structures and loss of depth information, making spatial relationships difficult to infer. This raises a fundamental question: how can we enable VLMs to reason effectively about 3D anatomy when only X-ray images are available? One possible approach is to reconstruct CT volumes from X-ray projections and subsequently perform 3D reasoning using a CT VLM. However, such reconstruction-based pipelines introduce additional computational overhead and potential error propagation. 

Rather than explicitly reconstructing CT, we explore transferring CT-informed reasoning directly from a CT teacher to an X-ray student. Recent studies have explored cross-modal on-policy distillation (OPD) to transfer reasoning capabilities across different input modalities~\cite{cao2026xopd, bousselham2026vold}. However, CT-to-X-ray reasoning transfer presents a distinct challenge due to the asymmetry in visual information between the teacher and student. The CT teacher has access to rich 3D anatomical information unavailable to the X-ray student, and directly applying teacher supervision may conflate CT-specific information with X-ray observations, hindering effective knowledge transfer. Therefore, effective transfer requires selectively distilling CT-informed reasoning into the student's own reasoning trajectories while preserving X-ray observations.

To address this challenge, we propose Dreaming in CT (DiCT), a progressive CT-to-X-ray reasoning transfer framework that enables X-ray VLMs to reason with CT-informed anatomical knowledge. DiCT consists of three complementary stages. First, instruct the dreamer employs structured supervised fine-tuning (SFT) to initialize CT-oriented reasoning and establish a structured interface with X-ray observations and CT-oriented reasoning, preparing the student for selective knowledge transfer. Building on this structure, visualize the dream introduces CT-guided OPD, where the CT teacher provides token-level guidance on the X-ray student's own reasoning trajectories using paired CT volumes. Specifically, teacher supervision is selectively applied to CT-oriented reasoning and final answers, while X-ray observations are preserved through regularization toward the SFT reference. Finally, manifest the dream adapts the distilled reasoning capabilities from digitally reconstructed radiographs (DRRs) to real clinical X-rays through group-relative policy optimization (GRPO), with KL regularization to retain the acquired knowledge. Together, these stages enable the X-ray student to observe in 2D and reason with CT-informed knowledge, without requiring explicit CT reconstruction at inference time.

We evaluate DiCT on Qwen3-VL-4B and Hulu-4B using the 3D-RAD benchmark~\cite{3drad}. Specifically, we pair Digitally Reconstructed Radiographs (DRRs) from DRR-RATE~\cite{hou2024drrrate} with the corresponding CT volumes from CT-RATE~\cite{hamamci2024ctchat}. We further evaluate DiCT on real clinical X-rays from ReXVQA~\cite{rexvqa}. As shown in Fig.~\ref{fig:ct_imagination}(a) and (b), DiCT effectively transfers CT-informed reasoning capabilities to X-ray models, achieving performance comparable to or better than CT reconstruction-based methods at substantially lower inference cost. Results on ReXVQA also show that DiCT performs well on real clinical X-rays. Ablation studies highlight the key role of CT-guided OPD in reasoning transfer and the complementary benefits of structured SFT and clinical GRPO. Our contributions are as follows:

\begin{itemize}[leftmargin=*,noitemsep,topsep=2pt]
\item We propose Dreaming in CT (DiCT), a three-stage progressive CT-to-X-ray reasoning transfer framework comprising instruct the dreamer, visualize the dream, and manifest the dream, which respectively initialize structured reasoning, transfer CT-informed knowledge, and adapt to real clinical X-rays.
\item 
We introduce selective CT-guided on-policy distillation to effectively transfer 3D anatomical reasoning capabilities from CT teachers to X-ray students under asymmetric visual evidence.
\item Extensive experiments demonstrate the effectiveness of CT-informed reasoning transfer, promising generalization to real clinical X-rays, and competitive performance with substantially lower inference cost than CT reconstruction-based approaches.
\end{itemize}

\section{Related Work}

\subsection{Medical Vision-Language Models}

The success of large vision-language models~\cite{li2023llava,liu2024visual,chen2024internvl} has inspired their adaptation to medical imaging, with early work focusing on modality-specific models. LLaVA-Med~\cite{li2024llava} adapted instruction-following VLMs to biomedical image understanding, while CheXagent~\cite{chen2024chexagent} and CXR-LLaVA~\cite{you2024cxr} demonstrated strong capabilities in chest radiograph interpretation and report generation. For volumetric CT imaging, CT-CHAT~\cite{hamamci2024ctchat} developed a foundation model for 3D CT volumes with language understanding capabilities. More recently, BTB3D~\cite{zhang2024btb3d} has advanced vision-language modeling in 3D medical imaging through improved tokenization. Generalist medical VLMs such as ClinFusion~\cite{clinfusion} and Hulu-Med~\cite{hulumed} are trained on mixed 2D and 3D datasets and can process multiple imaging modalities within unified architectures. However, these models emphasize modality coverage and image understanding, with limited attention to the relationship between paired CT volumes and X-ray projections.

\subsection{Cross-Modal Learning in Medical Imaging}

Cross-modal learning in medical imaging has explored various approaches to leveraging complementary information across imaging modalities. One prominent direction is X-ray-to-CT reconstruction, where methods such as X2CT-GAN~\cite{ying2019x2ct} and TF-PRDiT~\cite{tfprdit} reconstruct plausible 3D CT volumes from sparse X-ray projections. While these methods recover volumetric representations, they incur additional computational overhead at inference time and may introduce reconstruction errors. An alternative direction is cross-modal knowledge distillation, which transfers knowledge from a richer imaging modality to a model operating on a more accessible modality. For instance, Cahan et al.~\cite{cahan2025cxrctpa} explored CT-to-CXR knowledge distillation by transferring diagnostic representations from a CTPA teacher to a CXR student for pulmonary embolism diagnosis. However, existing approaches primarily focus on explicit CT reconstruction or task-specific knowledge transfer. Transferring structured reasoning from a CT teacher to an X-ray student remains largely unexplored.

\subsection{On-Policy Distillation}

Related optimization methods such as RLHF~\cite{ouyang2022training} and DPO~\cite{rafailov2024direct} align model behavior using preference signals, while PPO~\cite{schulman2017ppo} and GRPO~\cite{shao2024deepseekmath} support reward-driven policy optimization. OPD~\cite{agarwal2024onpolicy,gu2024minillm} provides dense teacher supervision on trajectories generated by the student, reducing the mismatch between training and inference associated with distilling fixed teacher outputs. Med-OPD~\cite{qian2026medopd} extends this paradigm to medical VLMs through evidence-aware supervision. Beyond medical imaging, X-OPD~\cite{cao2026xopd} aligns speech-conditioned students with text-based teachers using semantically matched inputs and token-level feedback. However, CT-to-X-ray reasoning transfer involves a distinct information asymmetry. The CT teacher observes volumetric evidence unavailable to the X-ray student. DiCT applies CT-guided OPD in this asymmetric setting to transfer reasoning capabilities from a CT teacher to an X-ray student.
\section{Method}

\begin{figure*}[t]
\centering
\includegraphics[width=\textwidth]{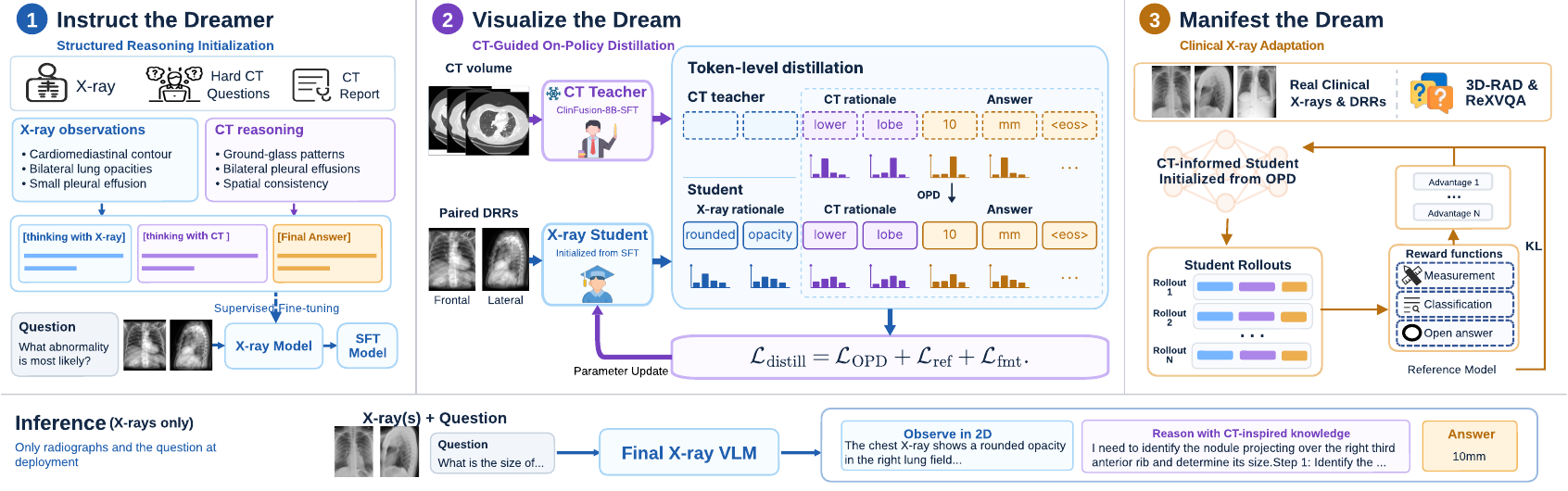}
\caption{\textbf{Method overview.} DiCT comprises three complementary stages: (1) instruct the dreamer initializes X-ray and CT-oriented reasoning through structured SFT. (2) visualize the dream applies masked CT-guided OPD from the CT teacher to the X-ray student's own responses. (3) manifest the dream adapts the distilled policy to real clinical X-rays through clinical GRPO with task rewards and a KL anchor. DiCT enables X-ray-only inference without CT reconstruction.}
\label{fig:method_overview}
\end{figure*}

We propose DiCT, a progressive CT-to-X-ray reasoning transfer framework that teaches X-ray students to observe in 2D and reason from an imagined CT perspective. Paired CT provides privileged volumetric guidance during training, enabling the student to reason as if it could see the corresponding CT volume while using only X-rays at inference. Figure~\ref{fig:method_overview} illustrates three complementary stages. In Sec.~\ref{sec:sft}, instruct the dreamer establishes a structured X-ray and CT reasoning interface. In Sec.~\ref{sec:cmpd}, visualize the dream transfers CT-informed knowledge through on-policy teacher guidance. And in Sec.~\ref{sec:grpo}, manifest the dream adapts the resulting policy from DRRs to real clinical X-rays. These stages address reasoning initialization, CT-to-X-ray reasoning transfer, and clinical adaptation, enabling X-ray-only inference without CT reconstruction.

\subsection{Problem Formulation}

Let $\mathcal{D}_{\text{X-ray}} = \{(x_i, q_i, a_i)\}_{i=1}^N$ denote X-ray images $x_i$, questions $q_i$, and reference answers $a_i$. We aim to train a vision-language policy $\pi_\theta$ generating $y \sim \pi_\theta(\cdot | x, q)$ while benefiting from volumetric supervision available only during training, without reconstructing CT at inference.

We assume a CT teacher $\pi_T$ and paired data $\mathcal{D}_{\text{paired}} = \{(x_i, c_i, q_i, a_i)\}$, where each X-ray $x_i$ corresponds to a CT volume $c_i$ of the same patient. The teacher observes $c$ while the student observes $x$. Our goal is to endow the X-ray student with CT-informed reasoning, enabling it to think beyond planar projections as if the corresponding CT were available. This asymmetry in visual evidence motivates the three-stage transfer framework.

\subsection{Instruct the Dreamer}
\label{sec:sft}

Instruct the dreamer prepares the X-ray student for CT-guided OPD through structured SFT. To enable selective CT teacher supervision, the student needs a consistent response structure that distinguishes X-ray observations from CT-oriented reasoning. We therefore initialize the student with structured chain-of-thought (CoT) responses that combine X-ray observations and CT-oriented reasoning, while using only X-rays as visual input.

\noindent\textbf{Structured reasoning template.}
To establish this separation, we organize each response into three parts.
\begin{itemize}[leftmargin=*,noitemsep,topsep=2pt]
\item Thinking with X-ray: Observations and reasoning based on 2D X-ray projections, including visible structures, intensity patterns, and planar relationships.
\item Thinking with CT: CT-oriented reasoning about anatomy, spatial relationships, and quantitative findings, grounded in the corresponding CT report.
\item Answer: The final result of the question.
\end{itemize}
This structured format provides the reasoning interface for selective teacher guidance in Sec.~\ref{sec:cmpd}. For comparison, we also construct an X-ray SFT baseline on the same training set, which generates X-ray reasoning followed by a final answer. Both settings use X-rays as visual input.

\noindent\textbf{Synthetic data generation.}
We use Qwen3.8-27B~\cite{qwen38} to synthesize reasoning chains across five radiology tasks: image observation, abnormality detection, Measurement, four-choice Abnormality Identification, and Longitudinal Diagnosis based on prior findings. For X-ray reasoning, we generate multiple CoT candidates conditioned on the X-ray images and questions, and apply rejection sampling to retain responses that pass format and task-specific answer verification checks. In parallel, we generate CT-oriented CoT from the corresponding CT reports and questions. The retained X-ray reasoning and report-grounded CT reasoning are then combined with the reference answer to form structured SFT targets. These examples constitute the training dataset $\mathcal{D}_{\mathrm{SFT}}$.

\noindent\textbf{SFT objective.}
We fine-tune the student on the structured reasoning responses using token-level cross-entropy:
\begin{equation}
\mathcal{L}_{\text{SFT}}=-\mathbb{E}_{(x,q,y)\sim\mathcal{D}_{\mathrm{SFT}}}\sum_t\log\pi_\theta(y_t\mid x,q,y_{<t}).
\end{equation}
This structured initialization serves two purposes. First, it exposes the X-ray student to CT-oriented anatomical reasoning derived from paired CT reports, providing an initial foundation for volumetric reasoning. Second, it establishes distinct reasoning sections that enable selective supervision in the subsequent distillation stage. However, SFT relies on fixed synthesized reasoning targets and cannot adapt its guidance to the student's evolving reasoning trajectories.

\subsection{Visualize the Dream}
\label{sec:cmpd}

Building on this initialization, visualize the dream introduces CT-guided OPD~\cite{agarwal2024onpolicy,gu2024minillm} to transfer knowledge through student-generated reasoning trajectories rather than fixed CoT targets. Specifically, the X-ray student generates its own responses, while a frozen CT teacher evaluates these trajectories token by token using the paired CT volume. This enables the teacher to provide guidance on the student's current reasoning behavior, allowing CT-informed knowledge transfer to extend beyond static supervision.

\noindent\textbf{Teacher-student asymmetry.}
Unlike conventional OPD settings, the CT teacher $\pi_T$ observes $c$, whereas the X-ray student $\pi_\theta$ observes $x$. Although the teacher can evaluate student-generated tokens using the same textual prefix, the two models condition on different visual evidence. Consequently, uniformly applying CT teacher supervision may encourage the student to attribute CT-specific findings to its X-ray observations, even when they are not directly supported by the projection. This motivates selective distillation that transfers CT-informed reasoning while preserving X-ray observations.

\noindent\textbf{Structured masking for selective distillation.}
We leverage the reasoning structure established in Sec.~\ref{sec:sft} to assign distinct supervision signals to different response sections. Specifically, we divide the generated tokens into three groups:
\begin{itemize}[leftmargin=*,noitemsep,topsep=2pt]
\item $M_{\text{X-ray}}$: Tokens in the ``Thinking with X-ray'' section, where reasoning should remain grounded in the student's 2D observations.
\item $M_{\text{CT}}$: Tokens in the ``Thinking with CT'' and ``Answer'' sections, where teacher guidance transfers CT-informed reasoning and answer-level knowledge.
\item $M_{\text{format}}$: Structural tokens, including section headers and delimiters, which maintain the shared reasoning format.
\end{itemize}
This separation allows the student to acquire CT-informed knowledge without directly imposing CT teacher supervision on X-ray observations.

\noindent\textbf{Cross-modal token advantage.}
For a response $y$ sampled from the student, we define the token-level advantage as the difference between the CT teacher and previous student log-probabilities:
\begin{equation}
A_t=\log\pi_T(y_t\mid c,q,y_{<t})-\log\pi_{\theta_{\mathrm{old}}}(y_t\mid x,q,y_{<t}).
\end{equation}
Since the teacher and student condition on different visual inputs, $A_t$ captures the teacher's relative preference for student-generated tokens rather than aligning visual representations. Let $r_t=\pi_\theta(y_t\mid x,q,y_{<t})/\pi_{\theta_{\mathrm{old}}}(y_t\mid x,q,y_{<t})$ and $\bar r_t=\operatorname{clip}(r_t,1-\epsilon,1+\epsilon)$. We apply teacher guidance only to CT-oriented reasoning and answer tokens through the masked OPD objective:
\begin{equation}
\mathcal{L}_{\text{OPD}}=-\lambda_T\,\mathbb{E}\sum_t M_{\text{CT}}^{(t)}\min(r_tA_t,\bar r_tA_t).
\end{equation}

To preserve X-ray reasoning, we instead regularize the student toward the frozen SFT reference on X-ray tokens:
\begin{equation}
\mathcal{L}_{\mathrm{ref}}=\lambda_{\mathrm{ref}}\,\mathbb{E}\sum_tM_{\mathrm{X-ray}}^{(t)}D_{\mathrm{KL}}(\pi_\theta\Vert\pi_{\mathrm{SFT}}).
\end{equation}
To maintain the structured reasoning format, we additionally replay structured SFT examples and apply supervised cross-entropy only to format tokens, while masking out reasoning and answer content:
\begin{equation}
\mathcal{L}_{\mathrm{fmt}}=-\lambda_{\mathrm{fmt}}\,\mathbb{E}\sum_tM_{\mathrm{format}}^{(t)}\log\pi_\theta(y_t^*\mid x,q,y_{<t}^*).
\end{equation}
The overall distillation objective combines three terms:
\begin{equation}
\mathcal{L}_{\mathrm{distill}}=\mathcal{L}_{\mathrm{OPD}}+\mathcal{L}_{\mathrm{ref}}+\mathcal{L}_{\mathrm{fmt}}.
\end{equation}
Together, these objectives selectively transfer CT-informed reasoning while preserving X-ray observations and maintaining the response structure, preparing the student for subsequent clinical adaptation.

\subsection{Manifest the Dream}
\label{sec:grpo}

While the preceding stages transfer CT-informed reasoning through paired CT volumes and DRRs, the distribution gap between DRRs and real clinical X-rays may limit generalization. To bridge this gap, manifest the dream adapts the distilled X-ray student through clinical GRPO~\cite{shao2024deepseekmath}. We train on ReXVQA Measurement questions from real clinical X-rays, mixed with a smaller portion of 3D-RAD questions to maintain reasoning diversity during adaptation. 

\noindent\textbf{Task-specific rewards.}
For each training question, we sample $K$ responses and assign task-specific rewards. For 3D-RAD, open-ended observation and abnormality-description answers receive a semantic correctness score $s\in\{1,\ldots,10\}$ from an LLM judge, normalized as $R_{\mathrm{open}}=(s-1)/9$. For measurements, we use an exponential reward based on the mean normalized error across matched numerical components after unit conversion. For a single length measurement with reference $m$ and prediction $\hat m$ in millimeters, the reward is $R_{\mathrm{meas}}=\exp(-|\hat m-m|/\max(5,0.2m))$, accounting for both absolute and relative errors. Multiple-choice questions receive binary correctness rewards. ReXVQA length measurements follow the same reward formulation. Invalid or truncated responses receive zero reward.

\noindent\textbf{Group-relative policy optimization.}
For each group of $K$ responses, we normalize the rewards to obtain the group-relative advantage:
\begin{equation}
\hat A^{(k)}=\frac{R^{(k)}-\mu_R}{\sigma_R+\delta},
\end{equation}
where $\mu_R$ and $\sigma_R$ denote the group mean and standard deviation, and $\delta$ is a stabilizing constant. Let $r_{kt}$ denote the current-to-old token probability ratio and $\bar r_{kt}=\operatorname{clip}(r_{kt},1-\epsilon_G,1+\epsilon_G)$. The GRPO objective is
\begin{equation}
\mathcal{L}_{\mathrm{GRPO}}
=-\mathbb{E}\sum_{k,t}
\min\left(r_{kt}\hat A^{(k)},
\bar r_{kt}\hat A^{(k)}\right).
\end{equation}

While GRPO adapts the student to task-specific feedback, unconstrained optimization may cause it to drift from the CT-informed reasoning acquired during distillation. We therefore use the frozen OPD checkpoint as a reference policy and introduce a KL regularization term:
\begin{equation}
\mathcal{L}_{\mathrm{adapt}}
=\mathcal{L}_{\mathrm{GRPO}}
+\lambda_{\mathrm{KL}}
D_{\mathrm{KL}}(\pi_\theta\Vert\pi_{\mathrm{OPD}}).
\end{equation}
This objective encourages clinical adaptation while discouraging drift from the distilled policy, enabling X-ray-only inference without CT reconstruction.

\section{Experiments}
\begin{table*}[tp]
\centering
\caption{\textbf{Results on 3D-RAD.} Acc./Matched are percentages. Obs./Abn. are observation/abnormality Judge scores (1--10). MAE (mm) uses matched length components. TFLOPs include reconstruction. Bold/underlined: best/second-best, excluding CT-input references.}
\label{tab:3drad_summary}
\begingroup
\small
\setlength{\tabcolsep}{2pt}
\renewcommand{\arraystretch}{0.95}
\begin{tabular*}{\textwidth}{@{\extracolsep{\fill}}l*{7}{c}@{}}
\toprule
\multirow[c]{2}{*}{Method} & \begin{tabular}[c]{@{}c@{}}Abnormality\\Identification\end{tabular}
& \begin{tabular}[c]{@{}c@{}}Longitudinal\\Diagnosis\end{tabular}
& \multicolumn{2}{c}{\begin{tabular}[c]{@{}c@{}}Open-ended\\Interpretation\end{tabular}}
& \multicolumn{2}{c}{Measurement} & Inference \\
\cmidrule(lr){2-2}\cmidrule(lr){3-3}\cmidrule(lr){4-5}\cmidrule(lr){6-7}\cmidrule(lr){8-8}
& Acc.$\uparrow$
& Acc.$\uparrow$ & Obs.$\uparrow$ & Abn.$\uparrow$
& MAE$\downarrow$ & Matched$\uparrow$ & TFLOPs$\downarrow$ \\
\midrule
\multicolumn{8}{@{}l}{\colorbox{blue!10}{\textit{I. Paired CT input: CT-input references}}} \\
ClinFusion-8B-SFT (CT teacher) & 56.07 & 79.74 & 5.32 & 4.87 & 8.50 & 94.5 & 22.17 \\
OmniCT-3B& 21.34 & 71.35 & 4.50 & 4.44 & 12.16 & 99.5 & 12.41 \\
OmniCT-7B& 22.12 & 78.35 & 4.41 & 4.56 & 11.20 & 98.1 & 19.82 \\
\midrule
\multicolumn{8}{@{}l}{\colorbox{violet!10}{\textit{II. X-ray $\rightarrow$ reconstructed CT $\rightarrow$ CT VLM}}} \\
TF-PRDiT + ClinFusion-8B-SFT & \underline{45.87} & 78.28 & \textbf{5.00} & \textbf{4.73} & \textbf{8.92} & 95.2 & 132.66 \\
X2CT-GAN + ClinFusion-8B-SFT & 42.20 & 77.97 & 4.96 & 4.56 & \underline{9.48} & 95.7 & 22.69 \\
\midrule
\multicolumn{8}{@{}l}{\colorbox{red!10}{\textit{III. Direct X-ray inference}}} \\
ClinFusion-8B-SFT & 36.56 & 79.50 & 4.88 & 4.40 & 9.87 & 96.3 & 23.31 \\
HealthGPT-Pro-4B& 27.73 & 71.32 & 4.57 & 4.24 & 14.87 & 100.0 & 5.94 \\
Hulu-4B & 26.17 & 64.46 & 4.70 & 4.08 & 12.40 & 100.0 & 7.38 \\
Hulu-4B (X-ray SFT $\rightarrow$ GRPO) & 41.37 & 79.50 & 4.57 & 4.08 & 14.22 & 100.0 & 7.24 \\
Hulu-4B-DiCT(ours) & \textbf{47.53} & \underline{79.60} & \underline{4.97} & \underline{4.58} & 9.86 & 100.0 & 10.39 \\
Qwen3-VL-4B & 18.09 & 65.37 & 3.53 & 3.13 & 15.77 & 76.9 & 11.75 \\
Qwen3-VL-4B (X-ray SFT $\rightarrow$ GRPO) & 34.69 & \textbf{80.06} & 4.41 & 4.03 & 12.22 & 100.0 & 7.01 \\
Qwen3-VL-4B-DiCT(ours) & 38.36 & 78.70 & 4.60 & 4.32 & 10.99 & 100.0 & 10.44 \\
\bottomrule
\end{tabular*}
\endgroup
\end{table*}

\subsection{Experimental Setup}

\paragraph{Baseline Methods.}
We evaluate DiCT on two VLM backbones, Qwen3-VL-4B~\cite{qwen3vl} and Hulu-4B~\cite{hulumed}. We use ClinFusion-8B-SFT, obtained by fine-tuning ClinFusion-8B~\cite{clinfusion}, as our CT teacher and CT-input baseline. We compare three inference paradigms: (1) direct CT reasoning with ClinFusion-8B-SFT and OmniCT-3B/7B~\cite{omnict}, (2) reconstruction-based reasoning using TF-PRDiT~\cite{tfprdit} or X2CT-GAN~\cite{ying2019x2ct} followed by the same CT teacher, and (3) X-ray-only reasoning with the teacher's native 2D branch, HealthGPT-Pro-4B~\cite{healthgpt}, the original X-ray student backbones, and DiCT. For a controlled comparison, we additionally evaluate X-ray SFT and X-ray SFT $\rightarrow$ GRPO baselines on the same X-ray student backbones, without CT-oriented reasoning or teacher distillation.

\paragraph{Datasets.}
We evaluate on 3D-RAD and ReXVQA, covering both controlled CT-to-X-ray reasoning transfer and generalization from DRRs to real clinical X-rays. For 3D-RAD~\cite{3drad}, we pair CT-based questions in CT-RATE~\cite{hamamci2024ctchat} with frontal and lateral DRRs from DRR-RATE~\cite{hou2024drrrate}, enabling evaluation across different input modalities on corresponding cases. We exclude temporal questions lacking prior imaging evidence. We also reformulate the original binary existence task, where 82.0\% of answers are negative, into a four-choice Abnormality Identification task. This reformulation addresses the severe class imbalance in the original task. The reformulated subset contains 4,223 questions across 1,134 volumes with approximately balanced answer positions. The resulting test set comprises 11,788 questions across five radiological tasks. Further construction details and results on the original binary task are provided in the Supp.~\ref{sec:task4_existence}.

ReXVQA~\cite{rexvqa} evaluates transfer to real clinical chest X-rays. We convert length questions based on images into open-ended numerical questions by removing multiple-choice options. This yields 4,027 training and 244 test questions. We additionally evaluate 1,547 native Abnormality Identification questions from the official test split. While similar identification tasks are included during structured SFT and CT-guided OPD using DRRs, ReXVQA Abnormality Identification data are not used for clinical GRPO. Subset construction is detailed in Supplementary.

\paragraph{Training Details.}

All stages use ms-swift, DeepSpeed ZeRO-3, and bf16 on four NVIDIA H20 GPUs, updating the language model while freezing the vision encoder and aligner. We obtain ClinFusion-8B-SFT by fine-tuning ClinFusion-8B on 30,662 CT question--answer pairs with reasoning data from Qwen3.8-27B~\cite{qwen38} for one epoch (480 updates), using a learning rate of $1\times10^{-5}$ and batch size 64. The resulting teacher is then frozen.

The X-ray student undergoes three one-epoch stages. Structured SFT uses 10,113 examples with a learning rate of $1\times10^{-5}$, followed by CT-guided OPD on 40,938 questions at $1\times10^{-6}$. OPD uses temperature 0.9, $\lambda_T{=}0.5$, $\lambda_{\mathrm{ref}}{=}0.02$, $\lambda_{\mathrm{fmt}}{=}0.1$, and $\epsilon{=}0.2$. Clinical GRPO uses the same learning rate, eight responses per question, and $\lambda_{\mathrm{KL}}{=}0.02$ on mixed ReXVQA Measurement and 3D-RAD questions. Task rewards follow Sec.~\ref{sec:grpo}, with invalid ReXVQA measurements receiving zero reward. ReXVQA Abnormality Identification is excluded from GRPO. The OPD checkpoint serves as both the GRPO initialization and frozen KL reference.

\subsection{Results on DRRs}
\label{sec:ct_to_xray_transfer}

\paragraph{Strong Performance across X-ray Models.}
Table~\ref{tab:3drad_summary} reports DiCT's performance across multiple radiological tasks on 3D-RAD using X-ray inputs alone. Compared with conventional X-ray SFT $\rightarrow$ GRPO, DiCT improves Abnormality Identification accuracy from 34.69\% to 38.36\% for Qwen3-VL-4B and from 41.37\% to 47.53\% for Hulu-4B. Both backbones also show improvements in open-ended interpretation and Measurement while maintaining comparable Longitudinal Diagnosis performance. These consistent gains across different backbones demonstrate the effectiveness of CT-informed reasoning transfer.

\paragraph{Comparison with CT Reconstruction.}
We further compare DiCT with pipelines that reconstruct CT volumes before reasoning. Hulu-4B-DiCT achieves 47.53\% Abnormality Identification accuracy, surpassing the 45.87\% achieved by TF-PRDiT + ClinFusion-8B-SFT, while reducing inference computation by 12.77$\times$. DiCT also outperforms the more efficient X2CT-GAN reconstruction pipeline in both accuracy and inference cost. Although reconstruction-based methods remain stronger on certain Measurement and open-ended interpretation tasks, DiCT provides a more favorable performance--efficiency trade-off. These results suggest that transferring CT-informed reasoning offers an efficient alternative to explicit volumetric reconstruction.

\begin{figure}[t]
\centering
\includegraphics[width=\columnwidth]{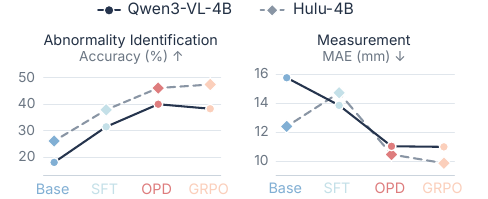}
\caption{\textbf{Training-stage effects on 3D-RAD.} Qwen3-VL-4B and Hulu-4B are compared across four training stages: Base, structured SFT, the OPD checkpoint, and DiCT after clinical GRPO. Complete results are provided in Supp.~\ref{sec:appendix_ablations}.}
\label{fig:dict_ablation}
\end{figure}

\paragraph{Contribution of CT-guided OPD.}
Figure~\ref{fig:dict_ablation} compares DiCT across successive training stages on 3D-RAD. Structured SFT establishes an initial X-ray/CT reasoning interface and improves Abnormality Identification, but does not consistently benefit Measurement. CT-guided OPD further improves Abnormality Identification accuracy from 31.54\% to 40.07\% for Qwen and from 37.96\% to 46.18\% for Hulu, while also reducing Measurement MAE. Table~\ref{tab:component_ablation_3drad} further examines the contributions of structured SFT and OPD on Hulu. The OPD-only baseline skips structured SFT and applies OPD to all generated tokens without selective masking. Combining structured SFT with OPD outperforms OPD alone on both metrics, demonstrating their complementarity. Moreover, the SFT+OPD checkpoint without clinical GRPO outperforms structured SFT followed by clinical GRPO without OPD. These findings highlight the central role of CT-guided OPD and the importance of structured initialization in effective reasoning transfer. They further support our central motivation that a structured reasoning interface allows privileged CT knowledge to guide the student's own X-ray-conditioned reasoning trajectories, without requiring CT access at inference time.

\subsection{Results on Real Clinical X-rays}
\label{sec:clinical_transfer}

\begin{table}[t]
\centering
\caption{\textbf{Results on ReXVQA.} Valid(\%) is the valid-answer rate. MAE(mm) uses valid responses.}
\label{tab:rexvqa}
\label{tab:rexvqa_identification}
\begingroup
\small
\setlength{\tabcolsep}{2.5pt}
\renewcommand{\arraystretch}{1.12}
\begin{tabular*}{\columnwidth}{@{\extracolsep{\fill}}lrrc@{}}
\toprule
\multirow[c]{2}{*}{Method} & \multicolumn{2}{c}{Measurement} & \multicolumn{1}{c}{\begin{tabular}[c]{@{}c@{}}Abnormality\\Identification\end{tabular}} \\
\cmidrule(lr){2-3}\cmidrule(lr){4-4}
& \shortstack[r]{Valid$\uparrow$} & \shortstack[r]{MAE$\downarrow$} & Acc.$\uparrow$ \\
\midrule
\multicolumn{4}{@{}l}{\textit{I. External X-ray references}} \\
ClinFusion-8B-SFT & 96.3 & 14.54 & 82.03 \\
HealthGPT-Pro-4B & 100.0 & 17.42 & 44.15 \\
\midrule
\multicolumn{4}{@{}l}{\textit{II. Qwen3-VL-4B}} \\
Base & 59.4 & 24.93 & 50.87 \\
X-ray SFT $\rightarrow$ GRPO & 100.0 & 14.35 & 35.42 \\
\textbf{DiCT} & 100.0 & 14.12 & 67.74 \\
\midrule
\multicolumn{4}{@{}l}{\textit{III. Hulu-4B}} \\
Base & 100.0 & 15.60 & 82.87 \\
X-ray SFT $\rightarrow$ GRPO & 100.0 & 13.43 & 75.11 \\
\textbf{DiCT} & 100.0 & 13.78 & 83.65 \\
\bottomrule
\end{tabular*}
\endgroup
\end{table}

\begin{table}[t]
\centering
\caption{\textbf{Training-component comparison on 3D-RAD.} \mbox{Hulu-4B} is evaluated on four-choice Abnormality Identification and Measurement. Complete results are provided in Supp.~\ref{sec:appendix_ablations}.}
\label{tab:component_ablation_3drad}
\begingroup
\small
\setlength{\tabcolsep}{3pt}
\renewcommand{\arraystretch}{1.12}
\newcommand{\componentyes}{\textcolor{teal!85!black}{$\checkmark$}}
\newcommand{\componentno}{\textcolor{red!75!black}{$\times$}}
\begin{tabular*}{\columnwidth}{@{\extracolsep{\fill}}*{5}{c}@{}}
\toprule
\multicolumn{3}{c}{Components} & \multicolumn{2}{c}{3D-RAD Performance} \\
\cmidrule(lr){1-3}\cmidrule(lr){4-5}
\begin{tabular}[c]{@{}c@{}}SFT\end{tabular}
& OPD
& \begin{tabular}[c]{@{}c@{}}GRPO\end{tabular}
& \begin{tabular}[c]{@{}c@{}}Abnormality\\Identification\\Acc. (\%)$\uparrow$\end{tabular}
& \begin{tabular}[c]{@{}c@{}}Measurement\\MAE (mm)$\downarrow$\end{tabular} \\
\midrule
\componentyes & \componentno & \componentyes & 38.34 & 12.23 \\
\componentno & \componentyes & \componentno & 40.94 & 11.40 \\
\componentyes & \componentyes & \componentno & \textbf{46.18} & \textbf{10.46} \\
\bottomrule
\end{tabular*}
\endgroup
\end{table}

\begin{figure}[t]
\centering
\includegraphics[width=\columnwidth]{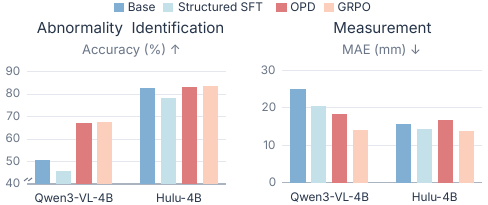}
\caption{\textbf{Training-stage effects on ReXVQA.} Abnormality Identification accuracy and Measurement MAE are evaluated for Qwen3-VL-4B and Hulu-4B across four stages: Base, structured SFT, the OPD checkpoint, and DiCT after clinical GRPO.}
\label{fig:rexvqa_ablation}
\end{figure}

\paragraph{Overall Performance on Real Clinical X-rays.}
Table~\ref{tab:rexvqa} compares DiCT with conventional X-ray training and external baselines on ReXVQA. On Abnormality Identification, DiCT achieves 67.74\% accuracy for Qwen3-VL-4B and 83.65\% for Hulu-4B, outperforming conventional X-ray SFT $\rightarrow$ GRPO on both backbones. On Measurement, DiCT remains competitive with conventional X-ray training, although it does not consistently outperform it. The improvements in Abnormality Identification across both backbones suggest that CT-informed reasoning remains beneficial when applied to real clinical radiographs rather than synthetic DRRs. Overall, DiCT demonstrates improved abnormality identification while maintaining competitive measurement accuracy on real clinical X-rays.

\paragraph{Generalization Before Clinical Adaptation.}
Figure~\ref{fig:rexvqa_ablation} shows that CT-guided OPD improves Abnormality Identification on real clinical X-rays even before clinical adaptation. Compared with structured SFT, accuracy increases from 45.64\% to 66.90\% for Qwen and from 78.15\% to 83.00\% for Hulu. The OPD checkpoint also surpasses the Qwen base model while remaining comparable to the already strong Hulu base model. Importantly, these improvements are achieved without training the OPD checkpoints on real clinical X-rays. These results suggest that CT-informed reasoning learned from paired DRRs and CT volumes can generalize to real clinical X-rays before clinical adaptation.

\paragraph{Clinical Adaptation and Capability Retention.}
Clinical GRPO further reduces ReXVQA Measurement MAE for both backbones, as shown in Figure~\ref{fig:rexvqa_ablation}. Since GRPO directly optimizes Measurement responses on real clinical X-rays, these improvements primarily reflect successful adaptation to the target domain. More importantly, we examine whether this adaptation affects the reasoning capabilities previously acquired through CT-guided OPD. ReXVQA Abnormality Identification examples are excluded from clinical GRPO, providing a separate evaluation of capability retention. Despite this, Abnormality Identification accuracy is maintained and slightly improved on both backbones. These results demonstrate that clinical GRPO effectively adapts the student to real clinical X-rays while preserving the CT-informed reasoning capabilities acquired through OPD, supporting the complementary roles of reasoning transfer and clinical adaptation in DiCT.

\subsection{Qualitative Analysis}
\label{sec:qualitative}

\begin{figure}[t]
\centering
\includegraphics[width=\columnwidth]{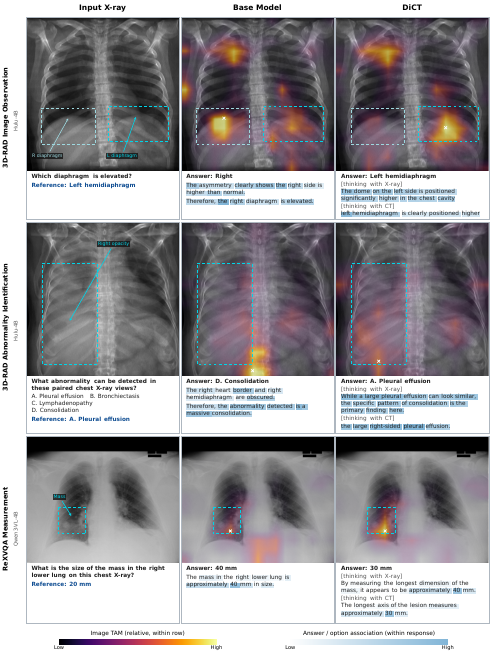}
\caption{\textbf{Qualitative improvements in visual focus and structured reasoning.} Rows show 3D-RAD Image Observation and Abnormality Identification (Hulu-4B), followed by ReXVQA Measurement (Qwen3-VL-4B). Columns compare input X-rays, base models, and DiCT. Token activation maps (TAM) target the dome, effusion, and mass, with shared scales within each row. Text scores are normalized per response. Cyan outlines indicate manually marked regions, and white crosses mark TAM maxima.}
\label{fig:tam_qualitative}
\end{figure}

Figure~\ref{fig:tam_qualitative} presents representative examples comparing DiCT with the base models, together with Token Activation Maps (TAM)~\cite{li2025tam} that visualize associations between generated text and image regions. The maps are obtained by replaying model responses on the original X-rays, averaging activations across occurrences of the target words. Text shading indicates associations with final-answer content, while multiple-choice responses are analyzed using content words from the selected options.

On 3D-RAD, DiCT correctly identifies the elevated left hemidiaphragm, whereas the base model incorrectly predicts the right. The corresponding activation shifts toward the relevant left diaphragmatic region. Similarly, in Abnormality Identification, the base model mistakes a prominent right-sided opacity for consolidation, while DiCT correctly identifies pleural effusion. Notably, although its initial X-ray reasoning favors consolidation, the imagined-CT reasoning revises the interpretation toward pleural effusion, accompanied by more focused visual associations near the right lung base. These examples illustrate how CT-informed reasoning helps resolve anatomical ambiguities in overlapping X-ray projections.

On ReXVQA Measurement, the base model estimates the mass size as 40\,mm. DiCT initially produces the same estimate from X-ray observations but refines it to 30\,mm through imagined-CT reasoning, halving the absolute error. The corresponding activation map shows a more concentrated response within the mass region. Together, these examples demonstrate that DiCT not only produces more accurate anatomical interpretations and quantitative estimates but also exhibits more anatomically aligned visual associations. This further highlights the effectiveness of transferring CT-informed reasoning to X-ray students for X-ray-only inference without CT reconstruction.

\section{Conclusion}
\label{sec:conclusion}

We introduced DiCT, a progressive framework for transferring CT-informed reasoning from CT teachers to X-ray vision-language students without explicit CT reconstruction at inference time. DiCT first establishes a structured interface separating X-ray observations from CT-oriented reasoning through SFT, then selectively transfers volumetric knowledge through CT-guided OPD, and finally adapts the acquired capabilities to real clinical X-rays through GRPO. Experiments across two VLM backbones on 3D-RAD and ReXVQA demonstrate improved X-ray-only reasoning and generalization to real clinical radiographs before adaptation. Subsequent GRPO improves measurement performance while maintaining abnormality identification accuracy. DiCT also achieves competitive performance against CT reconstruction-based approaches with substantially lower inference computation. Future work will explore extending CT-informed reasoning transfer to broader clinical settings and radiological tasks, enabling richer anatomical reasoning from accessible 2D radiographs.
\section*{LLM Usage Statement}
\label{sec:llm_usage}

We used GPT-6 to edit and polish the English prose of our paper. We used Qwen3.8-27B to synthesize CoT reasoning data for supervised fine-tuning of the CT teacher and X-ray students. We also used Qwen3.8-27B as a judge to score open-ended 3D-RAD responses for training rewards and evaluation. We designed the method, conducted the experiments, and interpreted the results. We reviewed the manuscript and take full responsibility for the entire submission, including all AI-assisted content and scientific conclusions.
{
    \small
    \bibliographystyle{ieeenat_fullname}
    \bibliography{main}
}

% Supplementary material is retained in this combined working draft.
% Remove the supplementary pages from the main-paper submission PDF.
\clearpage
\setcounter{page}{1}
\renewcommand{\thepage}{S\arabic{page}}
\maketitlesupplementary
\raggedbottom

\appendix
\begingroup
\iftoggle{cvprfinal}{}{\leftlinenumbers}
\section{Complete Ablation Results}
\label{sec:additional_results}
\label{sec:appendix_ablations}

\subsection{Complete Training-Stage Ablations}
\label{sec:full_ablation_results}

\endgroup

\begin{strip}
\begin{minipage}{\textwidth}
\captionsetup{type=table,position=top}
\centering
\caption{\textbf{Training-stage ablation on 3D-RAD.} Metrics and measurement protocol follow Table~\ref{tab:3drad_summary}.}
\label{tab:ablation_3drad}
\begingroup
\setlength{\tabcolsep}{2pt}
\renewcommand{\arraystretch}{1.05}
\resizebox{\linewidth}{!}{%
\begin{tabular}{@{}ll*{6}{c}@{}}
\toprule
\multirow{2}{*}{Backbone} & \multirow{2}{*}{Method} & Abnormality Identification
& Longitudinal Diagnosis
& \multicolumn{2}{c}{Open-ended Judge}
& \multicolumn{2}{c}{Measurement} \\
\cmidrule(lr){3-3}\cmidrule(lr){4-4}\cmidrule(lr){5-6}\cmidrule(lr){7-8}
& & Acc.$\uparrow$
& Acc.$\uparrow$ & Obs.$\uparrow$ & Abn.$\uparrow$
& MAE$\downarrow$ & Matched$\uparrow$ \\
\midrule
\multirow{6}{*}{\textit{Qwen3-VL-4B}} & Base & 18.09 & 65.37 & 3.53 & 3.13 & 15.77 & 76.9 \\
& structured SFT & 31.54 & \textbf{79.32} & \underline{4.60} & 4.11 & 13.86 & 99.9 \\
& SFT $\rightarrow$ GRPO & 33.08 & \underline{79.19} & 4.48 & 4.19 & 11.39 & 100.0 \\
& OPD w/o SFT & \underline{38.93} & 74.31 & 4.19 & 3.82 & 11.85 & 100.0 \\
& OPD checkpoint & \textbf{40.07} & 78.45 & \textbf{4.78} & \textbf{4.33} & \underline{11.03} & 100.0 \\
& DiCT & 38.36 & 78.70 & \underline{4.60} & \underline{4.32} & \textbf{10.99} & 100.0 \\
\midrule
\multirow{6}{*}{\textit{Hulu-4B}} & Base & 26.17 & 64.46 & 4.70 & 4.08 & 12.40 & 100.0 \\
& structured SFT & 37.96 & \underline{79.64} & 4.79 & 4.26 & 14.73 & 100.0 \\
& SFT $\rightarrow$ GRPO & 38.34 & 79.46 & 4.93 & 4.26 & 12.23 & 99.9 \\
& OPD w/o SFT & 40.94 & 73.27 & 4.58 & 4.06 & 11.40 & 100.0 \\
& OPD checkpoint & \underline{46.18} & \textbf{79.85} & \underline{4.95} & \underline{4.48} & \underline{10.46} & 99.8 \\
& DiCT & \textbf{47.53} & 79.60 & \textbf{4.97} & \textbf{4.58} & \textbf{9.86} & 100.0 \\
\bottomrule
\end{tabular}}
\endgroup
\end{minipage}
\end{strip}

Tables~\ref{tab:ablation_3drad} and~\ref{tab:ablation_rexvqa} report results after structured SFT, CT-guided OPD, and clinical GRPO for each backbone. Two controls vary the training sequence. The SFT $\rightarrow$ GRPO control applies clinical GRPO directly after structured SFT. The second applies standard OPD to the full response from the pretrained backbone. The latter control differs from CT-guided OPD in its initialization, the scope and weight of teacher supervision, X-ray reference regularization, and format replay (Sec.~\ref{sec:supp_distillation}).

From the OPD checkpoint to DiCT, ReXVQA Measurement MAE falls from 18.44 to 14.12\,mm for Qwen3-VL-4B and from 16.66 to 13.78\,mm for Hulu-4B. On 3D-RAD four-choice identification, accuracy changes from 40.07\% to 38.36\% for Qwen3-VL-4B and from 46.18\% to 47.53\% for Hulu-4B over the same stage.

\begin{table}[H]
\centering
\caption{\textbf{ReXVQA Measurement ablation and references.} Valid is the valid-answer rate (\%); MAE/RMSE are in mm over valid responses.}
\label{tab:ablation_rexvqa}
\begingroup
\small
\setlength{\tabcolsep}{2.5pt}
\renewcommand{\arraystretch}{1.12}
\begin{tabular*}{\columnwidth}{@{\extracolsep{\fill}}lccc@{}}
\toprule
Method & Valid$\uparrow$ & MAE$\downarrow$ & RMSE$\downarrow$ \\
\midrule
% External-reference RMSE values moved unchanged from the former main measurement table.
\multicolumn{4}{@{}l}{\textit{External X-ray references}} \\
ClinFusion-8B-SFT & 96.3 & 14.54 & 18.89 \\
HealthGPT-Pro-4B & 100.0 & 17.42 & 22.57 \\
\midrule
\multicolumn{4}{@{}l}{\textit{Qwen3-VL-4B}} \\
Base & 59.4 & 24.93 & 31.10 \\
structured SFT & 100.0 & 20.59 & 26.93 \\
SFT $\rightarrow$ GRPO & 100.0 & \textbf{14.03} & \textbf{17.94} \\
OPD w/o SFT & 100.0 & 20.57 & 26.22 \\
OPD checkpoint & 100.0 & 18.44 & 23.71 \\
DiCT & 100.0 & \underline{14.12} & \underline{17.97} \\
\midrule
\multicolumn{4}{@{}l}{\textit{Hulu-4B}} \\
Base & 100.0 & 15.60 & 20.62 \\
structured SFT & 100.0 & 14.45 & 18.85 \\
SFT $\rightarrow$ GRPO & 100.0 & \textbf{13.54} & \textbf{16.87} \\
OPD w/o SFT & 100.0 & 20.52 & 26.10 \\
OPD checkpoint & 100.0 & 16.66 & 21.67 \\
DiCT & 100.0 & \underline{13.78} & \underline{17.54} \\
\bottomrule
\end{tabular*}
\endgroup
\end{table}

\clearpage
\subsection{Complete ReXVQA Ablations}
\label{sec:rexvqa_ablations}

Clinical GRPO uses ReXVQA Measurement for adaptation, leaving ReXVQA Abnormality Identification as a separate test of transfer to clinical radiographs (Sec.~\ref{sec:supp_rexvqa}). On the same 1,547 identification questions, accuracy rises from 45.64\% after structured SFT to 66.90\% after OPD and 67.74\% after clinical GRPO for Qwen3-VL-4B. The corresponding accuracies for Hulu-4B are 78.15\%, 83.00\%, and 83.65\% (Table~\ref{tab:ablation_rexvqa_identification}).

\begin{table}[H]
\centering
\caption{\textbf{ReXVQA Abnormality Identification ablation.} Accuracy (\%) over 1,547 questions.}
\label{tab:ablation_rexvqa_identification}
\begingroup
\small
\setlength{\tabcolsep}{2.5pt}
\renewcommand{\arraystretch}{1.12}
\begin{tabular*}{\columnwidth}{@{\extracolsep{\fill}}lc@{}}
\toprule
Method & Acc.$\uparrow$ \\
\midrule
\multicolumn{2}{@{}l}{\textit{Qwen3-VL-4B}} \\
Base & 50.87 \\
structured SFT & 45.64 \\
SFT $\rightarrow$ GRPO & 51.13 \\
OPD w/o SFT & 63.48 \\
OPD checkpoint & \underline{66.90} \\
DiCT & \textbf{67.74} \\
\midrule
\multicolumn{2}{@{}l}{\textit{Hulu-4B}} \\
Base & 82.87 \\
structured SFT & 78.15 \\
SFT $\rightarrow$ GRPO & 78.86 \\
OPD w/o SFT & \underline{83.13} \\
OPD checkpoint & 83.00 \\
DiCT & \textbf{83.65} \\
\bottomrule
\end{tabular*}
\endgroup
\end{table}

\newpage
\subsection{Complete X-ray-Only Training Ablations}
\label{sec:xray_training_ablation}

\begin{table}[H]
\centering
\caption{\textbf{X-ray-only training on 3D-RAD.} Abn. ID and Long. denote Abnormality Identification and Longitudinal Diagnosis. Metrics and measurement protocol follow Table~\ref{tab:3drad_summary}.}
\label{tab:xray_training_3drad}
\begingroup
\setlength{\tabcolsep}{2pt}
\renewcommand{\arraystretch}{1.05}
\resizebox{\linewidth}{!}{%
\begin{tabular}{@{}l*{6}{c}@{}}
\toprule
\multirow{2}{*}{Method} & Abn. ID
& Long.
& \multicolumn{2}{c}{Judge}
& \multicolumn{2}{c}{Measurement} \\
\cmidrule(lr){2-2}\cmidrule(lr){3-3}\cmidrule(lr){4-5}\cmidrule(lr){6-7}
& Acc.$\uparrow$ & Acc.$\uparrow$
& Obs.$\uparrow$ & Abn.$\uparrow$ & MAE$\downarrow$ & Matched$\uparrow$ \\
\midrule
\multicolumn{7}{@{}l}{\textit{Qwen3-VL-4B}} \\
Base & 18.09 & 65.37 & 3.53 & 3.13 & \underline{15.77} & 76.9 \\
X-ray SFT & \underline{33.34} & \underline{79.46} & \textbf{4.53} & \underline{4.01} & 17.27 & 99.9 \\
X-ray SFT $\rightarrow$ GRPO & \textbf{34.69} & \textbf{80.06} & \underline{4.41} & \textbf{4.03} & \textbf{12.22} & 100.0 \\
\midrule
\multicolumn{7}{@{}l}{\textit{Hulu-4B}} \\
Base & 26.17 & 64.46 & \textbf{4.70} & \textbf{4.08} & \textbf{12.40} & 100.0 \\
X-ray SFT & \underline{38.86} & \underline{79.29} & 4.35 & \underline{4.04} & 20.32 & 95.5 \\
X-ray SFT $\rightarrow$ GRPO & \textbf{41.37} & \textbf{79.50} & \underline{4.57} & \textbf{4.08} & \underline{14.22} & 100.0 \\
\bottomrule
\end{tabular}}
\endgroup
\end{table}

Tables~\ref{tab:xray_training_3drad} and~\ref{tab:xray_training_rexvqa} report the X-ray-only controls. SFT uses an X-ray explanation and the reference answer as its target. GRPO then uses the same mixture of ReXVQA measurements and 3D-RAD questions as DiCT. With this training sequence, 3D-RAD four-choice accuracy reaches 34.69\% for Qwen3-VL-4B and 41.37\% for Hulu-4B, compared with 38.36\% and 47.53\%, respectively, for DiCT (Table~\ref{tab:ablation_3drad}).

\begin{table}[H]
\centering
\caption{\textbf{X-ray-only training on ReXVQA.} Metrics follow Table~\ref{tab:rexvqa}; RMSE is in mm over valid responses.}
\label{tab:xray_training_rexvqa}
\begingroup
\small
\setlength{\tabcolsep}{2.5pt}
\renewcommand{\arraystretch}{1.12}
\begin{tabular*}{\columnwidth}{@{\extracolsep{\fill}}lcccc@{}}
\toprule
\multirow[c]{2}{*}[2.4pt]{Method} & \multicolumn{3}{c}{Measurement}
& \multicolumn{1}{c}{\begin{tabular}[c]{@{}c@{}}Abnormality\\Identification\end{tabular}} \\
\cmidrule(lr){2-4}\cmidrule(lr){5-5}
& Valid$\uparrow$ & MAE$\downarrow$ & RMSE$\downarrow$ & Acc.$\uparrow$ \\
\midrule
\multicolumn{5}{@{}l}{\textit{Qwen3-VL-4B}} \\
Base & 59.4 & 24.93 & 31.10 & \textbf{50.87} \\
X-ray SFT & 100.0 & \underline{21.72} & \underline{27.11} & 31.29 \\
X-ray SFT $\rightarrow$ GRPO & 100.0 & \textbf{14.35} & \textbf{17.62} & \underline{35.42} \\
\midrule
\multicolumn{5}{@{}l}{\textit{Hulu-4B}} \\
Base & 100.0 & \underline{15.60} & \underline{20.62} & \textbf{82.87} \\
X-ray SFT & 100.0 & 17.20 & 22.19 & \underline{76.86} \\
X-ray SFT $\rightarrow$ GRPO & 100.0 & \textbf{13.43} & \textbf{16.17} & 75.11 \\
\bottomrule
\end{tabular*}
\endgroup
\end{table}

\clearpage
\flushbottom
\section{Binary Existence Detection}
\label{sec:task4_existence}

\subsection{Main Comparison}

The original 3D-RAD task asks whether each of 18 findings is present in each of 1,304 volumes. Among the 23,472 answers, 4,223 are positive and 19,249 are negative. An always-negative prediction thus yields 82.0\% accuracy and 50.0\% balanced accuracy. Table~\ref{tab:3drad_existence} reports both metrics for all three inference pathways. Accuracy measures the fraction of correct answers across all questions, while balanced accuracy gives equal weight to recall for each class. Section~\ref{sec:supp_3drad} describes how we use the same labels to construct the four-choice task.

\begin{table}[H]
\centering
\caption{\textbf{3D-RAD binary existence detection.} Acc./B.Acc.: accuracy/balanced accuracy (\%).}
\label{tab:3drad_existence}
\begingroup
\footnotesize
\setlength{\tabcolsep}{3pt}
\renewcommand{\arraystretch}{1.0}
\begin{tabular}{@{}lcc@{}}
\toprule
Method & Acc.$\uparrow$ & B.Acc.$\uparrow$ \\
\midrule
\multicolumn{3}{@{}l}{\textit{I. Paired CT input: CT-input references}} \\
ClinFusion-8B-SFT (CT teacher) & 81.16 & 64.86 \\
OmniCT-3B & 82.11 & 54.29 \\
OmniCT-7B & 82.50 & 54.84 \\
\midrule
\multicolumn{3}{@{}l}{\textit{II. X-ray $\rightarrow$ reconstructed CT $\rightarrow$ CT VLM}} \\
TF-PRDiT + ClinFusion-8B-SFT & 72.21 & \underline{61.01} \\
X2CT-GAN + ClinFusion-8B-SFT & 72.71 & 57.12 \\
\midrule
\multicolumn{3}{@{}l}{\textit{III. X-ray input}} \\
ClinFusion-8B-SFT & 78.36 & 56.24 \\
HealthGPT-Pro-4B & \textbf{82.17} & 51.25 \\
Hulu-4B & \underline{80.97} & 51.53 \\
Hulu-4B (X-ray SFT $\rightarrow$ GRPO) & 67.07 & 57.85 \\
Hulu-4B-DiCT(ours) & 78.07 & \textbf{61.86} \\
Qwen3-VL-4B & 78.70 & 49.78 \\
Qwen3-VL-4B (X-ray SFT $\rightarrow$ GRPO) & 70.04 & 52.02 \\
Qwen3-VL-4B-DiCT(ours) & 80.73 & 53.40 \\
\bottomrule
\end{tabular}
\endgroup
\end{table}

\subsection{Training-Stage Ablations}

Table~\ref{tab:ablation_3drad_existence} reports both metrics at each training stage. From the OPD checkpoint to DiCT, accuracy for Qwen3-VL-4B changes from 78.60\% to 80.73\% and balanced accuracy from 55.56\% to 53.40\%. The final checkpoint correctly identifies 453 of 4,223 positive findings and 18,495 of 19,249 negative findings. For Hulu-4B, accuracy changes from 79.69\% to 78.07\% and balanced accuracy from 59.03\% to 61.86\%.

\begin{table}[H]
\centering
\caption{\textbf{Training-stage ablation on binary existence detection.} Metrics follow Table~\ref{tab:3drad_existence}.}
\label{tab:ablation_3drad_existence}
\begingroup
\footnotesize
\setlength{\tabcolsep}{3pt}
\renewcommand{\arraystretch}{1.0}
\begin{tabular}{@{}lcc@{}}
\toprule
Method & Acc.$\uparrow$ & B.Acc.$\uparrow$ \\
\midrule
\multicolumn{3}{@{}l}{\textit{Qwen3-VL-4B}} \\
Base & \underline{78.70} & 49.78 \\
structured SFT & 75.63 & 52.57 \\
SFT $\rightarrow$ GRPO & 77.19 & \underline{53.42} \\
OPD checkpoint & 78.60 & \textbf{55.56} \\
DiCT & \textbf{80.73} & 53.40 \\
\midrule
\multicolumn{3}{@{}l}{\textit{Hulu-4B}} \\
Base & \textbf{80.97} & 51.53 \\
structured SFT & 75.35 & 57.91 \\
SFT $\rightarrow$ GRPO & 79.61 & 58.13 \\
OPD checkpoint & \underline{79.69} & \underline{59.03} \\
DiCT & 78.07 & \textbf{61.86} \\
\bottomrule
\end{tabular}
\endgroup
\end{table}

\subsection{X-ray-Only Training Ablations}

Under X-ray SFT followed by GRPO, balanced accuracy is 52.02\% for Qwen3-VL-4B and 57.85\% for Hulu-4B (Table~\ref{tab:xray_training_3drad_existence}), compared with 53.40\% and 61.86\%, respectively, for DiCT (Table~\ref{tab:3drad_existence}).

\begin{table}[H]
\centering
\caption{\textbf{X-ray-only training on binary existence detection.} Metrics follow Table~\ref{tab:3drad_existence}.}
\label{tab:xray_training_3drad_existence}
\begingroup
\footnotesize
\setlength{\tabcolsep}{3pt}
\renewcommand{\arraystretch}{1.0}
\begin{tabular}{@{}lcc@{}}
\toprule
Method & Acc.$\uparrow$ & B.Acc.$\uparrow$ \\
\midrule
\multicolumn{3}{@{}l}{\textit{Qwen3-VL-4B}} \\
Base & \textbf{78.70} & 49.78 \\
X-ray SFT & 70.00 & \textbf{52.71} \\
X-ray SFT $\rightarrow$ GRPO & \underline{70.04} & \underline{52.02} \\
\midrule
\multicolumn{3}{@{}l}{\textit{Hulu-4B}} \\
Base & \textbf{80.97} & 51.53 \\
X-ray SFT & \underline{69.93} & \underline{57.78} \\
X-ray SFT $\rightarrow$ GRPO & 67.07 & \textbf{57.85} \\
\bottomrule
\end{tabular}
\endgroup
\end{table}

\clearpage
\section{Benchmark Construction and Data Splits}
\label{sec:supp_benchmarks}

\subsection{3D-RAD}
\label{sec:supp_3drad}

For each CT-RATE volume in our 3D-RAD evaluation, the student receives the frontal and lateral DRRs and a question phrased for radiographs. The CT teacher receives the corresponding volume. The main test set comprises 11,788 questions across five tasks, with 1,024 on Image Observation, 2,666 on Abnormality Detection, 1,002 on Measurement, 4,223 on four-choice Abnormality Identification, and 2,873 on Longitudinal Diagnosis. Longitudinal Diagnosis provides the prior finding sequence in the question. We exclude the separate temporal task because it requires prior images that are unavailable in this setting.

\paragraph{Four-choice construction.}
We generate one question per positive finding, drawing three distinct distractors from findings labeled absent in the same volume. For example, a question about lung nodule uses medical material, bronchiectasis, and interlobular septal thickening as distractors. All 4,223 positive labels contribute a question, covering 1,134 volumes. The other 170 volumes contain only negative labels and enter the binary evaluation (Sec.~\ref{sec:task4_existence}). Positions A through D contain the correct answer 1,056, 1,056, 1,056, and 1,055 times, respectively. Balancing option positions preserves the observed finding frequencies. We use dataset labels independently of teacher predictions. We verified that the training and test sets share neither volume names nor image paths. Independent patient identifiers were unavailable for this check.

\subsection{ReXVQA}
\label{sec:supp_rexvqa}

\paragraph{Open-ended Measurement.}
We apply the same selection criteria to the official training, validation, and test splits (Table~\ref{tab:supp_rex_splits}). Starting from measurement questions with a physical quantity in the correct option, we retain those that specify a target and yield a single length convertible to millimeters. Distance questions must identify the reference landmark. We exclude questions that require another imaging modality or a prior examination. We remove the choices from the prompt and keep the converted value from the correct option as the scoring reference.

\begin{table}[H]
\centering
\caption{\textbf{ReXVQA Measurement subset construction.} Each column follows the same successive filters within its official split.}
\label{tab:supp_rex_splits}
\small
\begin{tabular*}{\columnwidth}{@{\extracolsep{\fill}}lrrr@{}}
\toprule
Selection step & Train & Val. & Test \\
\midrule
Numeric measurement candidates & 4,567 & 292 & 282 \\
Single length convertible to mm & 4,265 & 276 & 259 \\
Eligible open-ended questions & 4,027 & 262 & 244 \\
\bottomrule
\end{tabular*}
\end{table}

For example, we rewrite a question about endotracheal tube position to ask for the unsigned distance from the tube tip to the carina. The original answer, 3\,cm above the carina, gives a reference distance of 30\,mm. The measurement splits have disjoint patient and study identifiers. Of the 244 test questions, 223 ask for the distance from an airway tube tip to the carina, so this measurement accounts for most of the test set. We use the original answer choices as references throughout. The test set also includes one acromioclavicular joint measurement from a shoulder radiograph.

\paragraph{Abnormality Identification.}
From the official test split, we select four-choice questions that ask which finding is present on the current images. We require four distinct findings from a common vocabulary and preserve the wording, option order, and correct letter. We exclude normal or absent findings as targets, questions about change over time, and options differing only in location or severity. The official answer determines correctness. Findings unmentioned in the report remain unlabeled.

The 1,547 questions span 1,471 studies and 1,306 patients. We retain all associated images in their original order. This set shares neither patients nor studies with the 4,027 ReXVQA Measurement training questions. Identification therefore evaluates a distinct clinical task on patients outside the ReXVQA training subset used for GRPO.

\clearpage
\raggedbottom
\tcbset{suppprompt/.style={
  width=\linewidth, colback=gray!4!white, colframe=black!38,
  boxrule=0.45pt, arc=1.2mm, boxsep=0pt,
  left=8pt, right=8pt, top=6pt, bottom=6pt, middle=5pt,
  before skip=0pt, after skip=0pt,
  fontupper={\fontencoding{T1}\selectfont\small}, fontlower={\fontencoding{T1}\selectfont\small},
  before upper={\raggedright\setlength{\parindent}{0pt}},
  before lower={\raggedright\setlength{\parindent}{0pt}}
}}
\section{Training Data and Selective Distillation}
\label{sec:supp_distillation}

\subsection{Training sets and updates}

Table~\ref{tab:supp_training_sets} gives the number of training examples and optimizer updates at each stage. We fine-tune the CT teacher on CT question--answer pairs with synthesized reasoning, then freeze its weights. For student SFT, we generate candidate X-ray explanations from the paired DRRs and the question (Fig.~\ref{fig:supp_sft_prompts}a), retaining those that pass both format and task-specific answer checks. We assemble each structured target from the retained X-ray explanation, CT-oriented reasoning generated from the paired report (Fig.~\ref{fig:supp_sft_prompts}b), and the reference answer. The two DRRs are the student's only visual input. For the X-ray SFT control, the target consists of the retained X-ray explanation and the reference answer.
\begin{table}[H]
\centering
\caption{\textbf{Training data and optimizer updates.} Each question counts once regardless of the number of sampled responses. Counts exclude zero-loss batch padding.}
\label{tab:supp_training_sets}
\small
\begin{tabular*}{\linewidth}{@{\extracolsep{\fill}}lrr@{}}
\toprule
Training stage & Examples & Updates \\
\midrule
CT teacher SFT & 30,662 & 480 \\
Student structured SFT & 10,113 & 158 \\
CT-guided OPD & 40,938 & 427 \\
Clinical GRPO & 5,371 & 448 \\
\bottomrule
\end{tabular*}
\end{table}

The OPD set spans 12,896 CT volumes. Each OPD update uses 96 question slots and one student response per question. For clinical GRPO, we combine 4,027 ReXVQA Measurement questions with 1,344 3D-RAD questions covering all five main tasks. Each update allocates nine slots to ReXVQA and three to 3D-RAD, with eight sampled responses per question. Five padding entries complete the epoch and carry zero loss. The frozen OPD checkpoint serves as the KL reference throughout clinical GRPO.

% User-message templates from infer_eligible_xray_reasoning.build_prompt and generate_cot.build_user_prompt.
\clearpage
\begin{figure*}[t]
\centering
\begin{tcolorbox}[suppprompt]
\textsf{\bfseries (a) X-ray reasoning}\par{\footnotesize User message with frontal and lateral DRRs}\par\smallskip
\texttt{\{question\}}\par
Please reason step by step, and put your final answer within \texttt{\textbackslash boxed\{\}}.
\tcblower
\textsf{\bfseries (b) Report-grounded CT reasoning}\par{\footnotesize Text-only user message}\par\smallskip
You are tasked with generating a thinking process and a final answer for a given question based on a 3D chest CT volume.\par
\smallskip\textbf{Task Details:}\par
{\hangindent=1.3em\hangafter=1 1. The task model only sees the 3D chest CT volume and the question as its input and generates the thinking process and answer based on them.\par}
{\hangindent=1.3em\hangafter=1 2. You, however, are provided with the CT findings, CT impression, question, and correct answer as references for generating the thinking process and final answer.\par}
{\hangindent=1.3em\hangafter=1 3. Your response must simulate the task model's reasoning process, pretending that the findings are directly observed from the 3D chest CT volume. You must not reference or hint at the existence of the CT findings, CT impression, or correct answer provided as references.\par}
{\hangindent=1.3em\hangafter=1 4. The thinking process should outline step-by-step reasoning as if the observations are directly derived from the 3D chest CT volume. Avoid language or phrasing that implies knowledge from textual references.\par}
{\hangindent=1.3em\hangafter=1 5. Please put the final answer within \texttt{\textbackslash boxed\{\}}.\par}
{\hangindent=1.3em\hangafter=1 6. Begin directly with the image-grounded reasoning. Do not add a ``Thinking Process'' heading or meta-commentary about the user, prompt, task instructions, or how you will simulate the task model.\par}
\smallskip\textbf{Input Structure:}\par
{\hangindent=1.3em\hangafter=1 1. CT Findings: [The complete CT findings are provided for reference only. Do not reference or hint at them in your output.]\par}
{\hangindent=1.3em\hangafter=1 2. CT Impression: [The complete CT impression is provided for reference only. Do not reference or hint at it in your output.]\par}
{\hangindent=1.3em\hangafter=1 3. Question: [The specific question related to the 3D chest CT volume.]\par}
{\hangindent=1.3em\hangafter=1 4. Answer: [The correct answer, provided as ground truth for your guidance.]\par}
\smallskip\textbf{The input data is:}\par
CT Findings: \texttt{\{ct\_findings\}}\quad CT Impression: \texttt{\{ct\_impression\}}\par
Question: \texttt{\{question\}}\quad Answer: \texttt{\{reference\_answer\}}
\end{tcolorbox}
\caption{\textbf{Prompts for constructing structured SFT targets.} The panels show the templates for open-ended questions, with sample-specific inputs in braces. For multiple-choice questions, panel (a) appends the lettered options to the question. Panel (b) lists them under \emph{Answer Choices} and prefixes the reference answer with the correct letter when available.}
\label{fig:supp_sft_prompts}
\end{figure*}

\clearpage
\subsection{Supervision within a sampled response}

Figure~\ref{fig:supp_sampled_response} shows a student rollout. For teacher scoring, we append only the student's sampled CT reasoning and boxed answer to the teacher's question prompt for the paired volume. The student scores the same content in its full rollout context, including the preceding X-ray reasoning. This X-ray span receives KL regularization toward the frozen SFT reference. Section headers and delimiters receive format supervision on separate structured SFT examples.

% Training sample: Qwen3-VL-4B selective OPD, update 66, sample 8 (one-based).
% 3D-RAD training item: Task2_Anomaly_Detection/Abnormality_feature.csv:1068.
\begin{figure}[H]
\centering
\begin{tcolorbox}[
  width=\linewidth, colback=gray!5!white, colframe=blue!65!black,
  title={Qwen3-VL-4B during CT-guided OPD}, fonttitle=\bfseries,
  arc=3mm, boxrule=0.8pt, boxsep=0pt,
  left=8pt, right=8pt, top=5pt, bottom=6pt, toptitle=3pt, bottomtitle=3pt,
  before skip=0pt, after skip=0pt]
\small
\textbf{User prompt}\hfill{\footnotesize Frontal and lateral DRRs}\par
What is observed within the consolidation areas?\par
Please reason step by step, and put your final answer within \texttt{\textbackslash boxed\{\}}.
\tcblower
\small
\textbf{Student response}\par\smallskip
\texttt{[thinking with XRay]}\par\smallskip
\colorbox{orange!20}{\parbox[t]{\dimexpr\linewidth-2\fboxsep\relax}{\raggedright
Based on the chest X-ray images, particularly the frontal view, there are areas of increased opacity (whiteness) in the lung fields, which indicate consolidation. Within these consolidated areas, there are visible branching, tubular lucencies (dark areas). These represent air-filled bronchi that are more visible because the surrounding alveoli are filled with fluid or inflammatory exudate, a condition known as an air bronchogram. This is a classic sign of alveolar filling.}}
\par\smallskip
\texttt{[thinking with CT volume]}\par\smallskip
\colorbox{green!14}{\parbox[t]{\dimexpr\linewidth-2\fboxsep\relax}{\raggedright
Looking at the 3D chest CT volume, I can identify areas of increased attenuation (consolidation) in the lung parenchyma. Within these consolidated regions, I observe branching tubular structures that appear darker (air-density) against the surrounding opacified tissue. These structures represent air-filled bronchi that are visible through the consolidated alveolar spaces, which is a classic appearance of air bronchograms.}}\par\smallskip
\texttt{[Final Answer]}\quad
\colorbox{green!14}{\(\boxed{\text{Air bronchograms}}\)}
\end{tcolorbox}
\caption{\textbf{Selective supervision of a sampled response.} The student received only the two DRRs as visual input. Orange marks the X-ray reasoning regularized by the frozen SFT reference. Green marks the CT-oriented reasoning and answer content scored by the teacher using the paired CT. The final answer matches the reference label, \emph{Air bronchograms}.}
\label{fig:supp_sampled_response}
\end{figure}

\paragraph{Handling incomplete responses.}
The teacher loss is applied only when the sampled response contains both an identifiable CT reasoning span and a complete boxed answer. If either is missing, the X-ray reasoning span, when identifiable, can still contribute to the reference KL term.

\section{Evaluation Protocol and Inference Cost}
\label{sec:supp_evaluation}

\subsection{Answer extraction and scoring}

\paragraph{Choice and open-ended questions.}
For ReXVQA Abnormality Identification, both student backbones use greedy decoding with the original options and an instruction to box the final answer. We apply the same choice parser to every checkpoint and score all 1,547 questions. A response with no valid choice counts as incorrect. We also use choice accuracy for 3D-RAD Abnormality Identification and Longitudinal Diagnosis. On the original binary task, balanced accuracy is the arithmetic mean of positive and negative recall.

For 3D-RAD Image Observation and Abnormality Detection, the Qwen3.8-27B judge receives the question, reference answer, and candidate answer as text. It scores agreement on an integer scale from 1 to 10 using the messages in Fig.~\ref{fig:supp_judge_prompt}. Equivalent answers receive full credit. The rubric penalizes discrepancies in laterality, anatomical site, and finding identity. For evaluation, we average the raw scores within each task. For clinical GRPO, we convert a score \(s\) to a training reward of \((s-1)/9\).

\paragraph{3D-RAD Measurement.}
For the 3D-RAD Measurement tables, the prompt specifies the unit and answer form (a scalar, multiple dimensions, an interval, or a bound), without giving the target value. Of 1,002 questions, 1,000 request lengths and two request percentages. \emph{Matched} is the percentage of length questions with a parsable length of the required dimensionality. We convert centimeters to millimeters and treat unitless lengths as millimeters. We sort dimensions by magnitude before pairing them. For a scalar prediction and a reference interval, the error is zero inside the interval and the distance to the nearest endpoint outside it. We average absolute errors over all matched length components and score percentages separately.

\paragraph{ReXVQA Measurement.}
Each prompt requests one length in millimeters. For students, we parse the final boxed answer of a completed response. For direct-answer baselines, we parse the direct response. We accept one nonnegative value, converting centimeters to millimeters. Ranges, multiple values, and truncated responses are invalid. \emph{Valid} is the percentage of all 244 questions with a usable prediction. We compute MAE and RMSE over each model's valid predictions. For example, the Qwen3-VL-4B base model returns a valid length for 145 of 244 questions (59.4\%), so its errors are averaged over 145 predictions.

% System and user messages from the frozen 3D-RAD open-answer evaluation script.
\begin{figure*}[!t]
\centering
\begin{tcolorbox}[suppprompt]
\textsf{\bfseries System message}\par\smallskip
You are a strict, consistent evaluator of open-ended medical visual question answering. You do not see the medical image, so treat the reference answer as authoritative and judge only whether the candidate answer conveys the same medically relevant answer.\par
\smallskip\textbf{Score on this integer 1-10 scale:}\par
\noindent\makebox[1.6em][l]{\bfseries 10:}Fully correct, complete, and as precise as the reference; an exact or unambiguously equivalent answer.\par
\noindent\makebox[1.6em][l]{\bfseries 9:}Correct and equivalent, with only negligible wording or specificity differences.\par
\noindent\makebox[1.6em][l]{\bfseries 8:}Correct core answer with a small non-critical omission or imprecision.\par
\noindent\makebox[1.6em][l]{\bfseries 7:}Mostly correct, but has a meaningful minor omission, excess claim, or reduced specificity.\par
\noindent\makebox[1.6em][l]{\bfseries 6:}More correct than incorrect, but incomplete or partly imprecise.\par
\noindent\makebox[1.6em][l]{\bfseries 5:}Partially correct; captures roughly half of the essential answer.\par
\noindent\makebox[1.6em][l]{\bfseries 4:}Some relevant overlap, but misses or contradicts a major part of the answer.\par
\noindent\makebox[1.6em][l]{\bfseries 3:}Mostly incorrect, with only a small correct element.\par
\noindent\makebox[1.6em][l]{\bfseries 2:}Incorrect, but topically related or showing minimal relevant understanding.\par
\noindent\makebox[1.6em][l]{\bfseries 1:}Completely incorrect, contradictory, irrelevant, or no usable answer.\par
\smallskip\textbf{Rules:}\par
{\hangindent=0.8em\hangafter=1 - Evaluate semantic correctness, not exact string match, grammar, style, verbosity, or presence of reasoning.\par}
{\hangindent=0.8em\hangafter=1 - Do not reward unsupported details. Penalize a detail only when it changes, contradicts, or materially overstates the answer.\par}
{\hangindent=0.8em\hangafter=1 - For locations, laterality, anatomy, quantities, measurements, and named abnormalities, treat clinically meaningful mismatches as errors.\par}
{\hangindent=0.8em\hangafter=1 - For numerical answers, silently convert units and calculate the difference before scoring. Exact equivalent values can receive 10; a rounding difference up to 5\% can receive 9; an error up to 10\% can receive at most 8; an error from 10\% through 25\% can receive at most 7; larger errors require a lower score according to severity. These are upper bounds, not automatic scores.\par}
{\hangindent=0.8em\hangafter=1 - A concise answer can receive 10. Do not infer facts absent from both the reference and candidate.\par}
{\hangindent=0.8em\hangafter=1 - In the JSON, write the brief comparison reason first and the score second. Keep the reason under 160 characters.\par}
{\hangindent=0.8em\hangafter=1 - Return only JSON matching the requested schema.\par}
\tcblower
\textsf{\bfseries User message}\par\smallskip
Question: \texttt{\{question\}}\par
Reference answer: \texttt{\{reference\_answer\}}\par
Candidate answer: \texttt{\{candidate\_answer\}}\par
First compare the answers in one brief reason, then assign an integer score from 1 to 10.\par
\end{tcolorbox}
\caption{\textbf{Judge prompt for open-ended 3D-RAD answers.} The evaluator receives the question and the two answers as text. Its JSON response contains a brief \texttt{reason} and an integer \texttt{score} from 1 to 10.}
\label{fig:supp_judge_prompt}
\end{figure*}

\subsection{Counting inference operations}

We estimate the mean number of floating-point operations per question over \(N\) questions drawn from volumes \(\mathcal{V}\). For question \(i\), \(F_{\mathrm{VQA},i}\) includes visual encoding, prompt prefill, and decoding at the recorded input and output lengths, with key--value caching during decoding. All questions about volume \(v\) share one reconstruction, so its cost \(F_{\mathrm{recon},v}\) enters the total once. The mean cost per question is
\begin{equation*}
 \frac{1}{N}\left(\sum_{i=1}^{N} F_{\mathrm{VQA},i}
       +\sum_{v\in\mathcal{V}}F_{\mathrm{recon},v}\right).
\end{equation*}
\newpage
Direct X-ray and native CT inference have zero reconstruction cost. For TF-PRDiT and X2CT-GAN, we count the operations in the respective reconstruction pipelines and in the shared downstream CT teacher. We count two floating-point operations per multiply--add and include the dominant matrix, convolution, and attention operations. The estimates exclude CPU preprocessing, DRR synthesis, data transfer, batch padding, and the evaluation judge.

The reconstruction comparison uses 38,133 questions from 2,696 volumes. The X-ray SFT \(\rightarrow\) GRPO cost estimates use 35,260 questions, excluding the separate temporal task. We evaluate 3D-RAD task performance on the 11,788 questions in Sec.~\ref{sec:supp_3drad}.

\end{document}